\documentclass[letterpaper, 10 pt, conference]{ieeeconf}  

\IEEEoverridecommandlockouts                              

\usepackage{graphics} 
\usepackage{epsfig} 
\usepackage{amsmath} 
\usepackage{amssymb}  
\usepackage{bm}  
\usepackage[T1]{fontenc}

\title{\LARGE \bf
Pose Estimation of a Cable-Driven Serpentine Manipulator Utilizing Intrinsic Dynamics via Physical Reservoir Computing
}

\author{Kazutoshi Tanaka$^{1}$, Tomoya Takahashi$^{1}$, and Masashi Hamaya$^{1}$
\thanks{*This work was supported by JST, PRESTO Grant Number JPMJPR22C6, Japan.}
\thanks{$^{1}$K. Tanaka, T. Takahashi, and M. Hamaya are with OMRON SINIC X Corporation, Hongo 5-24-5, Bunkyo-ku, Tokyo, Japan
        {\tt\small kazutoshi.tanaka@sinicx.com}}%
}

\usepackage{fancyhdr}
\usepackage{url}

\begin{document}

\twocolumn[
\noindent
© 2025 IEEE. Personal use of this material is permitted. Permission from IEEE must be obtained for all other uses, in any current or future media, including reprinting/republishing this material for advertising or promotional purposes, creating new collective works, for resale or redistribution to servers or lists, or reuse of any copyrighted component of this work in other works.\\

\noindent
\textbf{Published article:}\\
K. Katayama, T. Takahashi, and M. Hamaya, ``Pose Estimation of a Cable-Driven Serpentine Manipulator Utilizing Intrinsic Dynamics via Physical Reservoir Computing,'' 2025 IEEE/RSJ International Conference on Intelligent Robots and Systems (IROS 2025), 2025.
]
\thispagestyle{empty}
\pagenumbering{gobble}
\clearpage

\maketitle
\thispagestyle{empty}

\begin{abstract}
Cable-driven serpentine manipulators hold great potential in unstructured environments, offering obstacle avoidance, multi-directional force application, and a lightweight design. 
By placing all motors and sensors at the base and employing plastic links, we can further reduce the arm's weight.
To demonstrate this concept, we developed a 9-degree-of-freedom cable-driven serpentine manipulator with an arm length of 545 mm and a total mass of only 308 g.
However, this design introduces flexibility-induced variations, such as cable slack, elongation, and link deformation. 
These variations result in discrepancies between analytical predictions and actual link positions, making pose estimation more challenging.
To address this challenge, we propose a physical reservoir computing based pose estimation method that exploits the manipulator's intrinsic nonlinear dynamics as a high-dimensional reservoir. 
Experimental results show a mean pose error of 4.3 mm using our method, compared to 4.4 mm with a baseline long short-term memory network and 39.5 mm with an analytical approach.
This work provides a new direction for control and perception strategies in lightweight cable-driven serpentine manipulators leveraging their intrinsic dynamics.
\end{abstract}

\section{INTRODUCTION}
\label{sec:introduction}

Cable-driven serpentine manipulators hold great potential in unstructured environments, offering obstacle avoidance, multi-directional force application, and a lightweight design.
They often incorporate redundant degrees of freedom (DOF), enabling versatile movement and adaptation to complex surroundings \cite{yoshikawa1985manipulability, maciejewski1985obstacle}.
Furthermore, a lightweight structure is crucial for safety and efficiency, as it reduces both energy consumption and the impact of unintended contact \cite{haddadin2007safety, haddadin2010soft, zanchettin2015safety}.

\begin{figure}[t]
  \centering
  \includegraphics[width=0.8\linewidth]{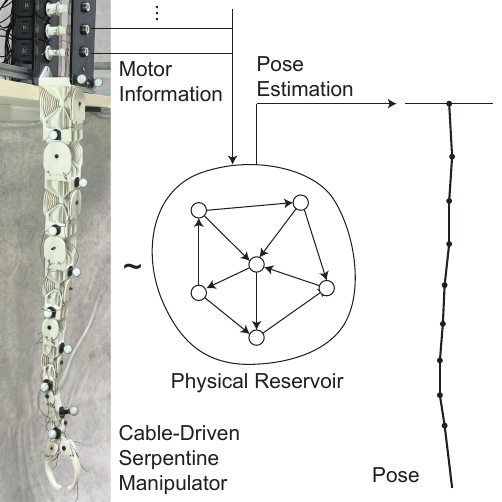}
  \caption{Pose estimation of a cable-driven serpentine manipulator actuated with motors and sensors at the base using physical reservoir computing.}
  \label{fig:teaser}
\end{figure}

Placing all motors and sensors at the base and using plastic links can further reduce the manipulator's arm weight while retaining robust force transmission.
Following this approach, we previously developed TwistSnake, a cable-driven serpentine manipulator with 6-DOF using plastic links~\cite{tanaka2023twist}.
Building on that design, we removed the pre-tension mechanism to reduce weight and introduced three additional DOFs, resulting in a 9-DOF cable-driven serpentine manipulator with an arm's length of 545~mm and its weight of only 308~g (Fig.~\ref{fig:teaser}).
Although these features facilitate highly flexible maneuvers and improve safety through lower energy consumption and reduced impact during unintended contact, they also introduce flexibility-induced variations, such as cable slack, cable elongation, and plastic link deformation.
These variations make pose estimation more challenging, causing discrepancies between analytical predictions and actual link positions.

To address this challenge, we propose a physical reservoir computing (PRC)~\cite{nakajima2020physical, tanaka2019recent} based pose estimation method that exploits the manipulator's intrinsic nonlinear dynamics as a computational resource.
Rather than treating the flexibility-induced variations as mere sources of uncertainty, we leverage them for computation within the PRC framework, thereby reducing the data and computational requirements typical of purely model-based or conventional learning-based approaches.
Experimental results show that our PRC-based method achieves a mean pose error of 4.3~mm, comparable to that of a standard long short-term memory (LSTM) network, while an analytical approach yields 39.5~mm.

Our primary contributions are as follows: 
\begin{itemize} 
    \item \textbf{Lightweight serpentine arm design:} A 9-DOF cable-driven serpentine manipulator with a 545~mm arm length and mass of only 308~g by removing the pre-tension mechanism and placing all motors and sensors at the base. 
    \item \textbf{Exploiting flexibility-induced variations:} A PRC-based pose estimation framework that capitalizes on the arm's intrinsic nonlinear dynamics, transforming what is typically problematic mechanical variation into a computational advantage. 
\end{itemize}

\section{RELATED WORK}
\label{sec:related}

\subsection{Pose estimation for manipulators without joint sensors}

Pose estimation for robotic manipulators without joint encoders has been pursued through a variety of approaches, including external vision systems~\cite{bohg2014robot}, inertial measurements~\cite{cheng2009joint}, and mechanism-intrinsic sensing~\cite{fennel2022calibration}. 
Bohg et al. proposed a pixel-wise part classification method for depth images, using a random forest to label each pixel by robot part and clustering those labels to infer 3D joint positions in real-time~\cite{bohg2014robot}. 
Cifuentes et al. introduced a probabilistic real-time tracking framework that combines depth camera data and joint sensing to account for sensor bias and calibration errors, achieving accurate manipulator pose estimation~\cite{cifuentes2016probabilistic}. 
Fennel et al. presented a calibration-free IMU-based approach employing an Extended Kalman Filter to fuse gyroscope and accelerometer data for kinematic state estimation without reliance on encoders~\cite{fennel2022calibration}. 
Labb{\'e} et al. developed a single-view pose estimation method, rendering a known 3D robot model and comparing it with an RGB image to infer both the end-effector pose and the joint angles~\cite{labbe2021single}. 
Kawaharazuka et al. proposed a technique for tendon-driven humanoids that uses only relative changes in muscle (tendon) lengths to estimate joint angles, thus accommodating complex musculoskeletal structures without extensive calibration~\cite{kawaharazuka2018method}.

In contrast to these sensor-rich or vision-based methods, we focus on a cable-driven serpentine manipulator without joint-embedded sensors, where pose estimation relies exclusively on the motors' angle and load. 
Moreover, we adopt a PRC framework that exploits the arm's intrinsic nonlinear dynamics for pose estimation.

Estimating the pose of flexible continuum arms has been widely explored, particularly in soft robotic systems where the absence of discrete joints makes direct measurement challenging. 
Many studies have investigated arms composed of soft materials (e.g., silicone rubbers) using embedded proprioceptive sensors to infer arm configuration, frequently applying machine learning techniques to relate sensor outputs to arm posture~\cite{chin2020machine, loo2021robust, thuruthel2019soft}. 
Soter et al. employed integrated bend sensors in an octopus-inspired soft robotic arm, pairing them with a stacked convolutional autoencoder (CAE) and an RNN for predictive visualization~\cite{soter2018bodily}, while Tariverdi et al. used RNNs to predict soft arm dynamics and facilitate closed-loop force and torque control~\cite{tariverdi2021}. 
Tanaka et al. adopted reservoir computing (RC) for pose estimation of a silicone-based soft arm, leveraging the material's inherent deformation~\cite{tanaka2022continuum}. 
Van Meerbeek et al. developed elastomer foam-based sensors with embedded optical fibers for proprioception, where machine-learning models classified deformation types and magnitudes through variations in light transmission~\cite{van2018soft}.

Although these works have significantly advanced proprioception in soft continuum arms, they do not address cable-driven serpentine manipulators with rigid segments and without joint-embedded sensors. 
In contrast, we maintain a lightweight structure by positioning all actuation and sensing at the base, using only motor angles and loads for estimation. 
Our approach leverages PRC to harness the intrinsic dynamics introduced by flexibility-induced variations of cable-driven serpentine manipulators, thereby enabling pose estimation inference without joint-embedded sensors.

\subsection{Physical reservoir computing for robotic state estimation}

PRC has been employed for a variety of robotic state estimation tasks, leveraging the system's intrinsic dynamics to process information. 
Bhovad and Li used an origami-based structure as a physical reservoir to perform nonlinear computations~\cite{bhovad2021physical}, while Calandra et al. employed PRC in quadruped robots to map proprioceptive leg joint data to exteroceptive conditions~\cite{calandra2021echo}. 
Kawase et al. introduced a pneumatic RC framework for estimating the posture of a soft exoskeleton by exploiting air flow within pneumatic tubes~\cite{kawase2021pneumatic}, and Horii et al. leveraged PRC for adaptive control in a soft swimming robot by analyzing its propulsion dynamics~\cite{horii2021physical}.

Despite these diverse robotic applications, none have focused on cable-driven serpentine manipulators, whose rigid-link structure and reliance on cables present unique challenges for pose estimation. 
Our work addresses this gap by applying PRC to a cable-driven serpentine manipulator, leveraging its intrinsic dynamics to enhance pose estimation. 
By centralizing sensing and actuation at the base, our approach maintains a lightweight arm design while providing pose estimation, distinguishing it from previous PRC applications in robotics.

\section{Cable-driven serpentine manipulator}

\subsection{Design}

\begin{figure}[t]
  \centering
  \includegraphics[width=\linewidth]{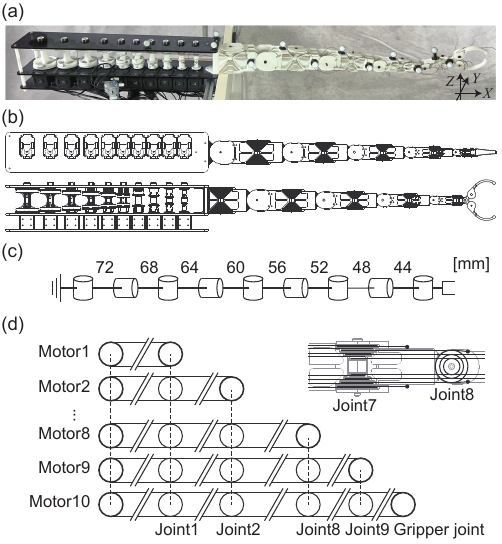}
  \caption{Our cable-driven serpentine manipulator. (a) Appearance and coordinates. (b) Design schematic. (c) Joint configuration. (d) Cable routing.}
  \label{fig:robot}
\end{figure}

Fig.~\ref{fig:robot} shows the overall appearance (a) and design schematic (b) of our cable-driven serpentine manipulator.
The arm itself is 545~mm long, while the total robot dimensions measure $969 \times 104 \times 80$~mm (L$\times$H$\times$D). 
By locating all motors (otherwise adding 57.2~g per joint, totaling 572~g if placed on links) and sensors at the base and removing joint-embedded sensors as in our previous Twist Snake~\cite{tanaka2023twist} design and removing the pre-tension mechanisms additionally, the moving links weigh only 308~g. 
The entire system weighs 1476~g.
Since the joints lack encoders, their angles are inferred from motor position data and an estimation model.

Fig.~\ref{fig:robot}(c) illustrates the joint configuration. 
The developed cable-driven serpentine manipulator has nine hinge joints in the arm, providing high kinematic redundancy for flexible movements, and a 1-DOF gripper. 
Each joint is rotated 90~degrees relative to the previous one along the movement direction. 

Unlike continuum arms made from flexible materials, this design choice contributes to its larger range of motion.
Each joint has a range of motion $[-180, 180]^\circ$, enabling extensive articulation in movement. 

The robot is actuated by ten servo motors (XL430-W250-R, ROBOTIS) mounted at the base link. 
The motors operate under velocity control, and they provide feedback on rotational angles and load derived from their current values. 
A 12~V power supply is used to drive the motors.

The motors drive polyethylene (PE) cables with a diameter of 4.05~mm and a load capacity of 30~kgf. 
The cable ends are fixed to the motor pulleys and links using friction-based fastening with screws.
The robot's structural components include 3D-printed nylon parts for the links and pulleys. 
The joint axes are constructed from 3~mm diameter machine-structural carbon steel rods, ensuring mechanical stability. 

Fig.~\ref{fig:robot}(d) depicts the cable routing. 
The cables pass through free rotational pulleys and route through the joints, implementing a coupled actuation system. 
This design ensures that when the cables transition from one joint to the next, they can move in a straight path without requiring guide pulleys.
In this configuration, moving a single motor influences multiple joints and all ten motors together actuate the arm's 10-DOF, leading to a fully actuated but coupling actuation mechanism. 
The thickness of the pulleys is 2.8~mm.
The radii of pulleys for motors 1 through 10 are 20, 18, ..., 4~mm, respectively. 
The groove distance between a pulley and its adjacent pulley is 2~mm, matching their radius difference between the adjacent motor pulleys.

\subsection{Control}

\begin{figure}[t]
  \centering
  \includegraphics[width=\linewidth]{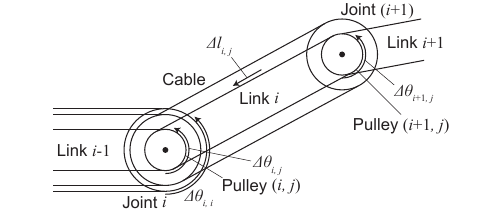}
  \caption{Rotational velocity of pulleys $\Delta \theta$ and cable velocity $\Delta l$ in adjacent two links.}
  \label{fig:pulley}
\end{figure}

The robot controls its pose, determining the target motor velocity from the desired joint velocity based on the relationship between motor and joint rotational velocity, as follows. 

Fig.~\ref{fig:pulley} illustrates the adjacent links, pulleys, and cables involved in actuation.
The rotational velocity of the pulley, located at the end of the $(i-1)^\mathrm{th}$ link and driven by the $j^\mathrm{th}$ motor, is given by $\Delta \theta_{i-1,j}$ and its corresponding radius is $r_{i-1,j}$.
Similarly, the rotational velocity of the pulley at the end of the $i^\mathrm{th}$ link is $\Delta \theta_{i,j}$ and its radius is $r_{i,j}$.
Assuming there is no slack or stretch in the cable and considering the relative rotational velocity to the $(i-1)^\mathrm{th}$ link, the relationship between the pulleys' rotational velocity, cable velocity $\Delta l_{i,j}$ and link rotational velocity $\Delta \theta_{i-1, i-1}$ follows:
\begin{eqnarray}
    r_{i, j} \Delta \theta_{i, j} &=& r_{i-1, j} (\Delta \theta_{i-1, j} - \Delta \theta_{i-1, i-1}) \\
    &=& \Delta l_{i,j}. \nonumber
\end{eqnarray}

For simplification, the robot design ensures that the radius of the pulleys through which the cable is routed and fixed is the same:
\begin{equation}
    r_{0, j} = r_{1, j} = ... = r_{j, j},  
\end{equation}
where $r_{0, j}$ represents the motor pulley radius.

Thus, the relationship between joint velocity $\Delta \theta_{1, 1}, ... , \Delta \theta_{9, 9}$ and motor velocity $\Delta \theta_{1, 0}, ... , \Delta \theta_{9, 0}$ is expressed as:
\begin{eqnarray}
    \Delta \theta_{1, 1} &=& \Delta \theta_{1, 0} \nonumber \\
    \Delta \theta_{2, 2} &=& \Delta \theta_{2, 0} - \Delta \theta_{1, 0} \nonumber \\
    &...& \nonumber \\
    \Delta \theta_{j, j} &=& \Delta \theta_{j, 0} - \Delta \theta_{j-1, 0}.
\end{eqnarray}
Based on these equations, the target motor velocity is calculated as 
\begin{equation}
    \Delta \theta_\mathrm{M} = A \Delta \theta_\mathrm{J}, 
    \label{eqn:vel}
\end{equation}
where $\theta_\mathrm{M}=[\Delta \theta_{1, 0}, ... , \Delta \theta_{9, 0}]^\top$ is a vector of motor velocity, 
$\theta_\mathrm{J}=[\Delta \theta_{1, 1}, ... , \Delta \theta_{9, 9}]^\top$ is a vector of joint velocity, 
and a coefficient matrix $A$ is 
\begin{equation}
A = 
    \left(
    \begin{array}{cccc}
      1   & 0   & ...   & 0 \\
      1   & 1   & ...   & 0 \\
      ... & ... & ...   & 0 \\
      1   & 1   & ...   & 1 
    \end{array}
        \right).
\end{equation}

In experiments, the robot controls its pose using Eq.~(\ref{eqn:vel}).
However, flexibility-induced variations, such as cable elongation and slack, in the robot result in a mismatch between the estimated joint angles using Eq.~(\ref{eqn:vel}) and actual ones.
Thus, our PRC-based method is required to estimate the arm's pose.

\section{METHOD}

This section first formulates the pose estimation task. 
Then, it provides an overview of PRC, illustrated in Fig.~\ref{fig:prc}. 
Last, we describe its application to pose estimation for cable-driven serpentine arms.

\subsection{Problem formulation for pose estimation}
We formulate the pose estimation task for the cable-driven serpentine manipulator introduced in the previous section. 
We represent the pose by the positions of optical motion-capture markers attached to each link. 
The markers' positions are measured with an external motion capture system during training, providing ground truth to capture the overall pose of the arm.
We aim to estimate the arm's pose, represented by the positions of optical markers attached to each link. 
We move the arm randomly, recording motor angles, motor load based on current, and the commanded motor velocities from the motors at the base as input signals. 
To estimate the pose, the robot learns a function $f_\mathrm{P}$ that maps these motor-based measurements $x$ to marker positions $y$, formally 
\begin{equation}
    y=f_\mathrm{P}(x).     
\end{equation}

During training, a model representing $f_\mathrm{P}$ is optimized so that the predicted marker positions match the measured ground-truth positions. 
The robot can estimate its pose using $f_\mathrm{P}$ without joint-embedded sensors.
We evaluate the performance of $f_\mathrm{P}$ by comparing the mean squared error (MSE) between the predicted and actual marker positions on a test dataset. 

\subsection{Physical reservoir computing}
\begin{figure}[t]
  \centering
  \includegraphics[width=\linewidth]{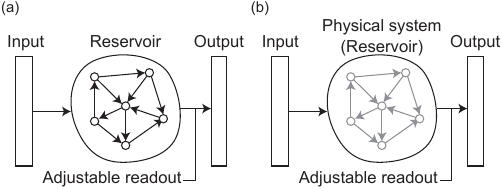}
    \caption{Physical reservoir computing (PRC). (a) Reservoir computing (RC), where the reservoir is a computational system. (b) PRC, where a physical system replaces the reservoir.}
  \label{fig:prc}
\end{figure}

Figure~\ref{fig:prc}(a) illustrates the basic concept of RC, while Fig.~\ref{fig:prc}(b) shows PRC, where the computational reservoir is replaced by a physical medium. 
PRC leverages the inherent dynamics of a physical system for efficient time-series processing. 
In this study, we exploit the nonlinear dynamics of our cable-driven serpentine manipulator, aiming to estimate its pose without requiring large datasets or extensive computation.

In RC (i.e., a computational neural network) and PRC  (i.e., a physical system), nodes evolve responses to input stimuli without requiring explicit training of internal parameters. 
Mathematically, let $ x_t \in \mathbb{R}^N $ be the reservoir state at time $t$, governed by a nonlinear dynamical system: 
\begin{equation} 
\frac{dx}{dt} = f(x_t, u_t), 
\end{equation} 
where $ u_t \in \mathbb{R}^M $ is the input signal, and $ f(\cdot) $ represents the physical system's governing function.

The readout function provides the final output of the reservoir.
In many works on PRC, the readout extracts relevant information via a linear transformation: 
\begin{equation} 
y_t = W_\mathrm{out} x_t, 
\end{equation} 
where $ y_t \in \mathbb{R}^K$ is a target variable and $W_{out} \in \mathbb{R}^{K \times N}$ is a trainable matrix optimized using standard regression techniques.

In contrast, inspired by previous studies~\cite{triefenbach2011can, masaad2023photonic}, we employ a nonlinear function as the readout 
\begin{equation} 
    y_t = f_\mathrm{out} (x_t) 
\end{equation} 
and train $f_\mathrm{out}$ using a multi-layer perceptron (MLP) network to represent complex relationships between $x$ and $y$. 
Although linear readout $W_\mathrm{out}$ can be trained from fewer data, learning $f_\mathrm{out}$ enables more accurate $y$ prediction.


Time multiplexing is a technique in RC that treats a single physical nonlinear node as multiple virtual nodes by splitting the input signal across distinct time intervals~\cite{nakajima2015information}.
This approach reduces hardware complexity and resource demands, as one component can emulate an extensive network by rapidly sampling its output in successive time frames.
In our implementation, we employed a simple form of time multiplexing known as delay vector mapping, where past system states are concatenated to form an extended input vector, to improve our PRC's performance as described in the following subsection.

\subsection{Pose estimation using physical reservoir computing}
\label{sec:method_c}
In this study, we apply PRC to our cable-driven serpentine manipulator by leveraging its complex, mechanically linked motion as a computational resource. 
Due to the arm's flexible structure, external inputs at the base propagate through multiple joints, creating intertwined dynamics that can serve as a high-dimensional reservoir. 
We hypothesize that this mechanical interference among joints improves PRC performance for pose estimation.

We define the state variables as follows. 
Let $s_t \in \mathbb{R}^{18}$ denote the concatenation of all motor angles and force measurements (one angle and one force per motor for nine motors to move the arm). 
Additionally, let $u_t \in \mathbb{R}^{9}$ (nine motors for the arm) represent the target motor velocities. 
Using time multiplexing with a window size $H$, we construct the input vector $ x_t = (s_t, s_{t-1}, ..., s_{t-H+1}, u_{t-1}, ..., u_{t-H+1} \in \mathbb{R}^{27H - 9}) $, where $u_t$ is not affecting $y_t$ and is not included in $x_t$ ($27H - 9 = 18 + 27 \times (H - 1)$).
The output signal is the 3D coordinates of markers placed on the cable-driven serpentine arm's links: $ y = (x_1, y_1, z_1, ..., x_9, y_9, z_9) \in \mathbb{R}^{27}$,  where $(x_j,\,y_j,\,z_j)$ represents the position of the $j^\mathrm{th}$ marker.
In our PRC framework, the vector $x_t$ represents the reservoir's high-dimensional state, while the readout function $f_\mathrm{out}$ is obtained by a multi-layer perceptron (MLP). 
We train $f_\mathrm{out}$ to map $x_t$ to $y_t$, thereby estimating the arm's pose from the system's intrinsic dynamics.

\section{EXPERIMENT}
We conducted a series of experiments to evaluate the proposed method from multiple perspectives. 
Specifically, we sought to address the following questions:
\begin{itemize}
    \item (Q1) Can the proposed method obtain improved accuracy compared to an analytical model in flexibility-induced variations?
    \item (Q2) How does the estimation accuracy of the proposed method compare to an LSTM baseline, as well as the ablation cases where a linear readout or force information is not used?
    \item (Q3) Does time series information contribute to improving estimation accuracy, and is there an appropriate sequence length for optimal performance?
\end{itemize}

\subsection{Data collection}
To evaluate the proposed pose estimation method, we placed the robot so that the $X$-axis of the coordinates in Fig.~\ref{fig:robot}(a) aligns with the direction of gravity and collected data by moving the serpentine arm randomly and measuring its pose, motor angles, and motor load.
Each link of the robot was equipped with an optical motion capture marker, and marker positions were tracked using a motion capture camera (V120 Trio, OptiTrack) at 120~Hz. 
The robot received the motors' angle and load data, as well as the marker positions, and sent the target velocity to the motors. 
It logged all measurements and the target, computed control commands, and finally sent these commands to the motors at 4~Hz.
To focus on motion without external forces and ensure movements remain within a collision-free range, the target joint angles were assigned within $(-22.5, 22.5)^\circ$ in joint 1, 3, 5, 7, and 9 and $(0, 22.5)^\circ$ in joint 2, 4, 6, and 8, respectively. 
The velocity command for each joint was determined using proportional (P) control, based on the difference between the target and current joint angles. 
The motor velocity was computed from the target joint velocity using Eq.~(\ref{eqn:vel}).
Target angles were updated every five steps.

We conducted 23 experimental sessions, each consisting of 2,100 control steps. 
For each session, the initial 100 steps were discarded to remove transient effects, leaving 2,000 usable steps per session. 
In total, this yielded 46,000 data points (2,000 steps $\times$ 23 sessions). 
Of these, sessions 1 through 20 (40,000 data points) were used for training, session 21 and session 22 for validation, and session 23 for final testing.

\subsection{Estimation algorithm}
For pose estimation, we trained the proposed PRC model (PRC-MLP) along an LSTM as a baseline model. 
The LSTM with one hidden layer was chosen as a standard recurrent network to capture sequential dependencies.
For LSTM, we replaced the current time step's target motor velocity---which does not affect the current pose---with zeros.
We also compared the proposed method (PRC-MLP) with a version that excludes load information (No-load) and another version with a linear layer as its readout (PRC-LIN).

In the proposed method, the physical reservoir state was generated by serializing the input data, and the output function using MLP was trained as a readout function as described in Sec.~\ref{sec:method_c}. 
All data were normalized in $(-1, 1)$. 
Models were trained using mini-batch learning, with hyperparameters, such as the learning rate and the number of nodes, optimized using validation data. 
The optimal sequence length ($H=4$) was first determined using the proposed method and then applied to LSTM and other baselines.

The models were trained using MSE loss, and pose estimation accuracy was evaluated based on the mean error of all markers.
The mini-batch size was set to 512. 
We used the Adam optimizer~\cite{kingma2014adam} for training. 
We employed a learning rate schedule starting at $10^{-3}$. After every epoch, we evaluated the model on the validation set. 
If the validation loss increased for three consecutive checks, the learning rate was reduced by a factor of 0.1. 
Early stopping was triggered once the learning rate dropped below $10^{-6}$.
MLP in the proposed method had four layers with a 512 layer size.
LSTM's hidden layer size was 512. 

To examine how the flexibility-induced variations affect pose estimation, we measured the error of an analytical approach using knowledge of marker locations on the links without filters (Analytical). 
At the test session in this condition, the initial joint angles were estimated from the measured initial marker positions based on a marker location on each link, then updated in sequence using Eq.~(\ref{eqn:vel}) to compute changes in joint angles from motor angle displacements. 
Based on these estimated joint angles, we calculated the positions of the markers. 

\subsection{Results}


\begin{figure*}[t]
  \centering
  \includegraphics[width=\linewidth]{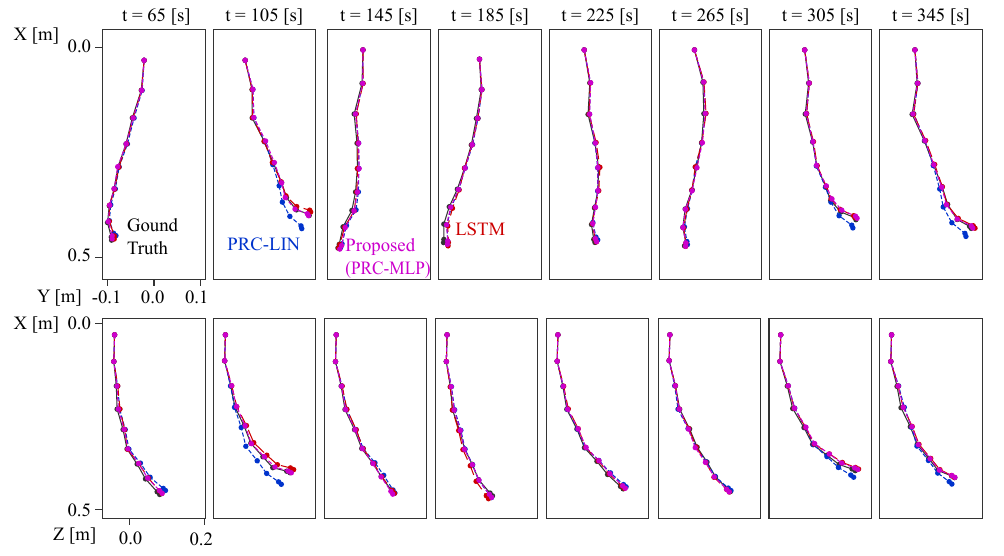}
  \caption{Snapshots of the serpentine manipulator's pose. The coordinates are in Fig.~\ref{fig:robot}(a).}
  \label{fig:snap}
\end{figure*}

\begin{figure}[t]
  \centering
  \includegraphics[width=\linewidth]{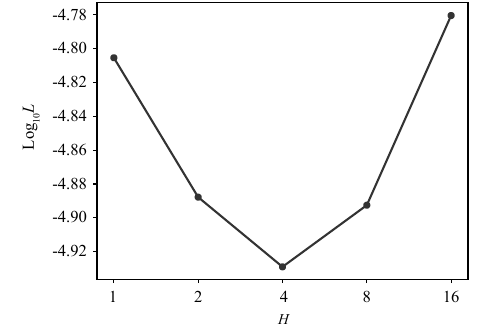}
    \caption{Prediction accuracy changes by sequence length. Validation loss $L$ with different sequence lengths $H$.}
  \label{fig:loss}
\end{figure}

\begin{table}[b]
    \caption{Estimation MSE. mean $\pm$ std [mm].}
    \label{tab:mse}
    \centering
    \begin{tabular}{|l|l|l|l|l|}
        \hline
        \textbf{ours}  & Analytical      & No-load       & PRC-LIN       & LSTM          \\
        \hline
        4.3 $\pm$ 3.6  & 39.5 $\pm$ 23.9 & 5.1 $\pm$ 4.3 & 7.2 $\pm$ 7.0 & 4.4 $\pm$ 3.6 \\
        \hline
    \end{tabular}
\end{table}

We investigated the estimation performance of the proposed method and other conditions.
Tab.~\ref{tab:mse} presents the marker position errors on the test set. 
The table shows that the proposed method demonstrates a prediction accuracy comparable to LSTM and outperforms No-load, PRC-LIN, and Analytical.
These results indicate that the proposed method and other learning methods effectively compensate for mechanical influences that contribute to estimation errors (Analytical).
Additionally, the results suggest that the force information (No-load) and the nonlinear function of the readout (PRC-LIN) in the proposed method improve the proposed method's estimation performance.
Moreover, the results demonstrate that the physical system of the cable-driven serpentine manipulator bears the computational burden of sequential processing in the proposed method, while LSTM processes sequential information in a computer. 


Fig.~\ref{fig:snap} illustrates the estimated poses predicted by the proposed method (PRC-MLP), PRC-LIN, and LSTM. 
In this figure, each pose is represented by the center points of the estimated joint positions, calculated from the marker positions.
The figure shows that the proposed method predicts poses similarly to LSTM and is more accurate than PRC-LIN.

We investigate whether the sequential information improves estimation performance and also how long a sequence length is required for pose estimation.
Fig.~\ref{fig:loss} represents loss $L$ with different sequence lengths $H$.
The $L$ decreases as the sequence length increases, reaches its minimum value with $H=4$, and increases.
These results indicate that sequential information improves performance.
Additionally, the results suggest that the proposed method requires a specific sequence length for pose estimation.
We presume that this is because PRC has enough nonlinearity and computational resources with such a sequence length, and overfitting occurs with a larger length than such a length.  

\section{DISCUSSION}
Our experimental results indicate that the proposed PRC approach results in a mean pose estimation error of 4.3~mm, which is nearly identical to the 4.4~mm error of a baseline LSTM network. 
The LSTM relies on computational resources to capture temporal features, whereas the PRC leverages the cable-driven serpentine manipulator's intrinsic nonlinear dynamics arising from flexibility-induced variations as a reservoir. 
Using MLP as the readout in our PRC approach benefits from reduced computational complexity and faster inference compared to LSTM, while also being well-suited for data-efficient learning and online pose estimation.
These findings highlight the potential for harnessing physical phenomena in robotic systems to mitigate the need for extensive computational resources and large datasets.

This work provides evidence that PRC can be successfully applied to pose estimation in the 9-DOF arm of the cable-driven serpentine manipulator with lightweight plastic links. 
By placing motors and sensors at the base, it becomes possible to reduce the weight of the moving links significantly. 
The proposed approach inherently accounts for the flexibility-induced variations, such as cable elongation, reducing discrepancies between analytical predictions and actual link positions. 
Consequently, this study presents a new example of using PRC for pose estimation in robotics~\cite{tanaka2021flapping} and paves the way for lightweight manipulator designs~\cite{tanaka2023twist}.
Our approach can also be applied to soft robot arms~\cite{thuruthel2019soft} and other cable-driven robots~\cite{kawaharazuka2018method}.

A primary limitation of this study is that our pose estimation relies on training data under an inference condition. 
Applying a previously acquired model in an online setting or zero-shot generalization across varied operating conditions and different manipulators~\cite{lu2024adaptive} remains an open challenge. 
Additionally, while we observed using a nonlinear readout can improve estimation accuracy in our experiments, it may reduce the data efficiency that purely linear readouts typically offer. 
Recent methods suggest incorporating nonlinear transformations in the readout while retaining linear layers for training~\cite{ohkubo2024reservoir}, which could enable both high accuracy and improved data efficiency.

Beyond pose estimation, leveraging PRC for end-effector force estimation~\cite{feng2021learning}, advanced control~\cite{tanaka2021trans}, and object state estimation~\cite{yu2022semi} in cable-driven serpentine manipulators is a promising direction for future work. 
Particularly, cable-driven serpentine manipulators offer a promising platform for learning object interactions through force application and estimating object poses by leveraging contact~\cite{von2020contact}, force feedback~\cite{nguyen2024symmetry, fuchioka2024robotic} and tactile feedback~\cite{murali2022active}. 
Such developments would expand the applicability of the serpentine manipulators using the PRC-based approach in tasks requiring fine manipulation or interaction with unstructured environments. 
Furthermore, as PRC's readout mechanisms and training strategies continue to evolve, integrating online learning and adaptive algorithms may lead to greater robustness and versatility in real-world deployments.

\section{CONCLUSIONS}
\label{sec:conclusion}

This study introduced a pose estimation method for cable-driven serpentine manipulators using PRC. 
By utilizing the inherent dynamics of the system, we demonstrated that our approach effectively estimates the pose of a cable-driven serpentine manipulator without requiring complex analytical models. 
Our experimental results confirmed that the proposed method obtained accuracy comparable to that of a standard LSTM.
This work provides a new direction for control and perception strategies in lightweight cable-driven serpentine manipulators leveraging their intrinsic dynamics.

\bibliographystyle{IEEEtran}
\bibliography{bib}

\end{document}